\documentclass[sigconf,natbib=true,anonymous=false]{acmart}

\usepackage{multirow}
\usepackage{xcolor}
\usepackage{placeins}
\usepackage{dblfloatfix}
\usepackage{balance}
\usepackage{booktabs}
\graphicspath{{./}{figs/}}

\AtBeginDocument{%
  }

\copyrightyear{2026}
\acmYear{2026}
\setcopyright{cc}
\setcctype{by}

\acmConference[CIKM '26]
{Proceedings of the 35th ACM International Conference on Information and Knowledge Management}
{November 07--11, 2026}
{Rome, Italy}

\acmBooktitle{Proceedings of the 35th ACM International Conference on Information and Knowledge Management (CIKM '26), November 07--11, 2026, Rome, Italy}

\acmDOI{10.1145/3799682.3840046}
\acmISBN{979-8-4007-2539-5/2026/11}

\begin{document}


\title{BERT4DTI : BERT-based Model for Predicting Drug-Protein Interactions}



\author{Thanina Hamitouch}
\affiliation{%
  \institution{Ecole Nationale Supérieure d'Informatique (ESI)}
  \city{Algiers}
  \country{Algeria}
}

\author{Khadidja Henni}
\affiliation{%
  \institution{Institut d'Intelligence Artificielle Appliquée, TELUQ University}
  \city{Montreal}
  \country{Canada}
}
\email{khadidja.henni@teluq.ca}

\author{Abdelkrim Aries}
\affiliation{%
  \institution{Ecole Nationale Supérieure d'Informatique (ESI)}
  \city{Algiers}
  \country{Algeria}
}
\email{ab\_aries@esi.dz}

\author{Amina Selma Haichour}
\affiliation{%
  \institution{Ecole Nationale Supérieure d'Informatique (ESI)}
  \city{Algiers}
  \country{Algeria}
}

\author{Neila Mezghani}
\affiliation{%
  \institution{Institut d'Intelligence Artificielle Appliquée, TELUQ University}
  \city{Montreal}
  \country{Canada}
}

\author{Lina Abou-Abbas}
\affiliation{%
  \department{Department of Electrical and Computer Engineering}
  \institution{Lebanese American University}
  \city{Byblos}
  \country{Lebanon}
}

\renewcommand{\shortauthors}{Thanina Hamitouch et al.}

\begin{abstract}
Understanding how drugs interact with protein targets is fundamental to drug discovery, drug repurposing and the early identification of promising therapeutic candidates before costly experimental testing. Sequence-based DTI models face three practical limitations: labelled interactions are scarce and unevenly distributed, large pretrained chemical and protein encoders are expensive to fine-tune end-to-end, and independently encoded sequences do not capture pair-specific dependencies. We present BERT4DTI, which encodes SMILES strings with ChemBERTa and amino-acid sequences with ProtBERT, applies bidirectional mutual attention between token-level representations, and classifies the resulting interaction features using convolutional layers and a multilayer perceptron. To reduce trainable size, ProtBERT is truncated to 18 retained layers and only the last two layers of each encoder are fine-tuned. On BIOSNAP, DAVIS and BindingDB, BERT4DTI is competitive, achieving the best ROC-AUC and PR-AUC on BIOSNAP and the highest sensitivity on all three benchmarks. An ablation on DAVIS shows that mutual attention improves PR-AUC and specificity. With 125M trainable parameters compared with 353M for full BERT fine-tuning, BERT4DTI provides a favourable performance--parameter trade-off for sequence-based DTI screening, while leaving runtime profiling, calibration and leakage-audited validation for future work.
\end{abstract}



\maketitle

\section{Introduction}
The therapeutic action of a drug depends strongly on its interactions with
biological targets, usually proteins. Identifying such drug--target interactions
(DTIs) is therefore central to drug discovery, off-target analysis and drug
repurposing~\cite{hughes2011principles,sachdev2019comprehensive}. Experimental
screening remains costly and time-consuming, making \emph{in silico} DTI
prediction useful for prioritizing plausible compound--protein pairs before
laboratory validation.

DTI prediction remains challenging. Known interactions form a small and biased
subset of the chemical--protein space, and many benchmarks construct negative
pairs from unknown interactions, introducing label uncertainty~\cite{ezzat2019computational,silva2026matrix}.
Moreover, drugs and protein targets are heterogeneous modalities with distinct
vocabularies, lengths and biological meanings. 

A model must therefore capture interaction-relevant information using sequence- or graph-based representations. Graph-based approaches model structural relationships~\cite{henni2025structural}, whereas others learn directly from drug and protein sequences~\cite{abouabbas2026novelty}.

Sequence-based deep learning methods avoid manual descriptor engineering.
DeepDTA~\cite{ozturk2018deepdta} uses CNNs over SMILES and amino-acid sequences,
MolTrans~\cite{huang2021moltrans} models substructure interactions with
Transformers, and attention-based models such as ICAN~\cite{kurata2022ican} and
HyperAttentionDTI~\cite{zhao2022hyperattentiondti} highlight the value of
cross-sequence alignment. In parallel, pretrained language models became strong
representation learners: ChemBERTa~\cite{chithrananda2020chemberta} learns from
large SMILES corpora, while ProtBERT from ProtTrans~\cite{elnaggar2022prottrans}
learns from protein sequences. Kang et al.~\cite{kang2022finetuning} combined
ChemBERTa and ProtBERT for DTI prediction, and DLM-DTI~\cite{lee2024dlmdti}
explored more economical adaptation through hint-based learning. Related
affinity-oriented models such as OdinDTA~\cite{xu2023odindta} and
ArkDTA~\cite{gim2023arkdta} further show the interest of attention mechanisms,
but they mainly target continuous affinity or interpretability objectives.
BERT4DTI focuses instead on binary sequence-based screening under common DTI
benchmarks.

Pretrained encoders improve representation quality but create a practical
trade-off. Full fine-tuning of large chemical and protein encoders requires many
trainable parameters, whereas frozen encoders may not adapt sufficiently to DTI
prediction. In addition, late concatenation of independently encoded drug and
protein embeddings does not explicitly model which tokens become relevant in
the paired sequence. This motivates a model that combines restricted
fine-tuning with explicit cross-modal interaction learning.

We propose BERT4DTI, a sequence-based DTI classifier built around
partially fine-tuned ChemBERTa and ProtBERT encoders, bidirectional mutual
attention, and a convolutional classification head. Our contributions are:
(i) a DTI architecture that aligns drug and protein token representations through
mutual attention; (ii) an evaluation on BIOSNAP, DAVIS and BindingDB with a
metric-level interpretation that separates sensitivity gains from PR-AUC and
specificity trade-offs, together with a component ablation; and (iii) a
trainable-parameter analysis and a cautious statement of the remaining limits
related to runtime, calibration, SMOTE/merged-data evidence and leakage-aware
validation.

\section{The BERT4DTI Method}
\subsection{Problem Definition and Inputs}
\begin{figure}[t]
  \centering
  \includegraphics[width=\columnwidth]{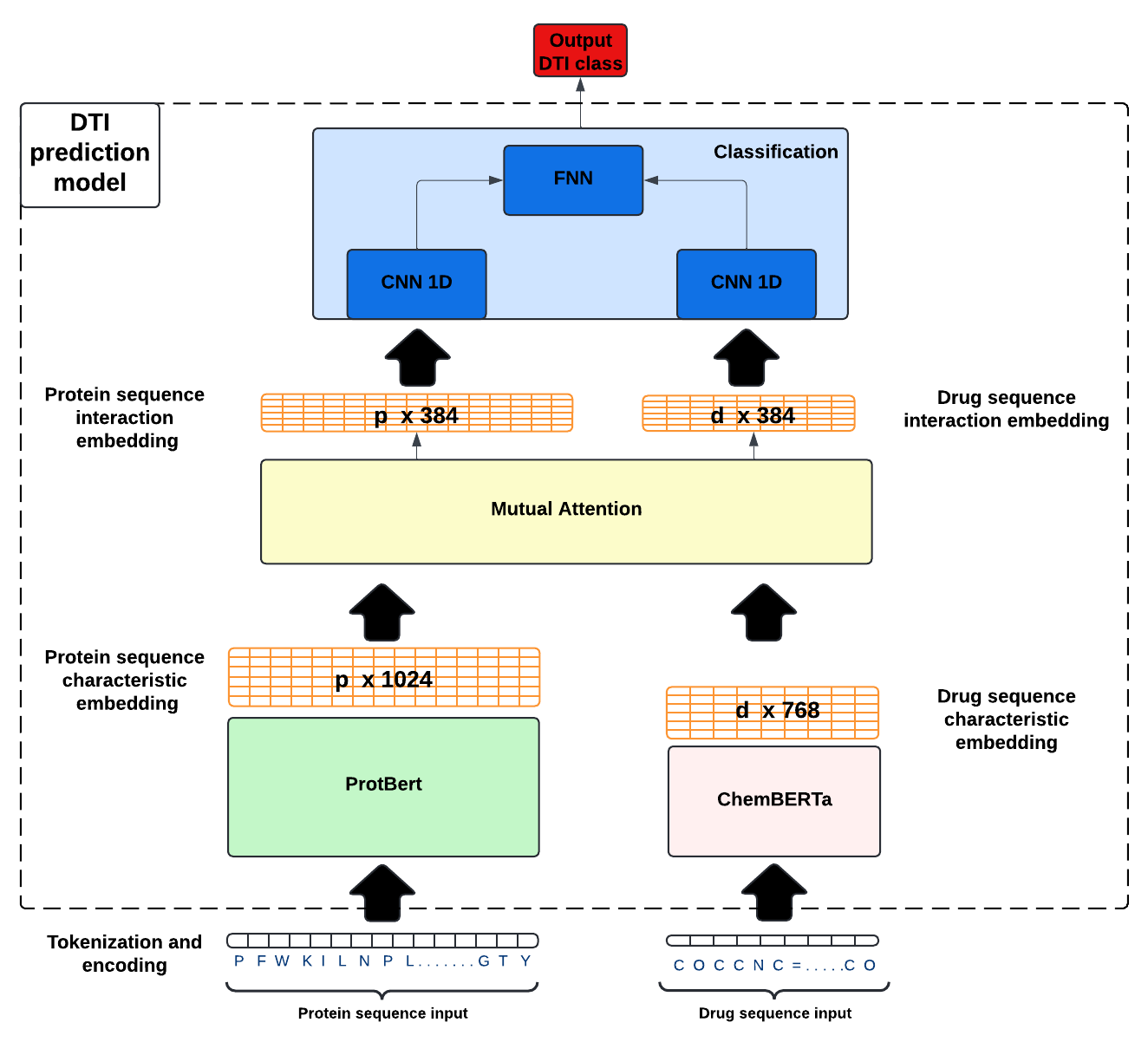}
  \caption{Overview of BERT4DTI. Contextual drug and protein token embeddings
    are mutually aligned before convolutional classification.}
  \Description{Block diagram: SMILES tokens enter ChemBERTa and amino-acid
    tokens enter ProtBERT; both feed a mutual-attention block; the resulting
    drug and protein interaction embeddings feed two 1D CNN branches whose
    outputs are concatenated and classified by a feedforward network into a
    binary DTI class.}
  \label{fig:arch}
\end{figure}

DTI prediction is a binary classification problem. Each example is a pair
$(d,p)$ where $d$ is a drug as a SMILES sequence and $p$ a protein as an
amino-acid sequence. The model learns $F:\mathcal{D}\times\mathcal{P}\to[0,1]$,
where $F(d,p)$ estimates the probability of interaction; a pair is positive
when this probability exceeds a validation-selected threshold. As shown in
Fig.~\ref{fig:arch}, the architecture has four stages: two pretrained encoders
produce contextual token representations; both modalities are projected to a
common dimension; a bidirectional mutual-attention block produces
interaction-conditioned representations; and independent 1D CNN blocks followed
by an MLP produce the prediction.
This binary setting follows the standard benchmark protocol used by MolTrans and pretrained-BERT baselines, enabling direct comparison with prior sequence-based DTI studies.

\subsection{Chemical and Protein Encoders}
Drugs are represented by canonical SMILES and encoded with ChemBERTa, a
RoBERTa-style model pretrained with masked language modelling on $\approx$10M
SMILES strings~\cite{chithrananda2020chemberta}:six Transformer layers, twelve
heads, hidden size 768, $\approx$83.4M parameters. Proteins are encoded with
ProtBERT from ProtTrans~\cite{elnaggar2022prottrans}; the full model has 30 layers and $\approx$420M parameters. In our configuration ProtBERT is reduced
to 18 retained layers ($\approx$268M parameters before freezing) to control the large memory cost of long protein sequences.

\paragraph{Partial fine-tuning.}
Let $\theta_d$ and $\theta_p$ denote ChemBERTa and the retained ProtBERT
parameters, partitioned into frozen and trainable parts
$\theta_d=(\theta_d^{\mathrm{fr}},\theta_d^{\mathrm{tr}})$,
$\theta_p=(\theta_p^{\mathrm{fr}},\theta_p^{\mathrm{tr}})$. Only the final two
layers of each retained encoder are updated, i.e.
$\nabla_{\theta_d^{\mathrm{fr}}}\mathcal{L}=0$ and
$\nabla_{\theta_p^{\mathrm{fr}}}\mathcal{L}=0$. This preserves most pretrained
parameters while allowing limited task adaptation; it is intended to reduce
trainable size, not to claim runtime gains without measured times. Encoder
outputs are linearly projected and normalized to a shared dimension $d_e$:
\begin{equation}
  D=\mathrm{LN}(E_dW_d+b_d),\qquad P=\mathrm{LN}(E_pW_p+b_p),
\end{equation}
with $D\in\mathbb{R}^{L_d\times d_e}$ and $P\in\mathbb{R}^{L_p\times d_e}$.

\subsection{Bidirectional Mutual Attention}
Late concatenation captures global information but does not ask which drug
tokens matter given a protein, or vice versa. BERT4DTI uses bidirectional
mutual attention. With $h$ heads, head $i$ uses projections
$Q^{D}_i=DW^{Q,D}_i,\,K^{D}_i=DW^{K,D}_i,\,V^{D}_i=DW^{V,D}_i$ for the drug and
$Q^{P}_i,K^{P}_i,V^{P}_i$ analogously for the protein. The
drug-conditioned-by-protein and protein-conditioned-by-drug outputs are
\begin{equation}
  \begin{aligned}
    H^{D}_i&=\mathrm{softmax}\!\Big(\tfrac{Q^{D}_i(K^{P}_i)^{\top}}{\sqrt{d_h}}\Big)V^{P}_i,\\
    H^{P}_i&=\mathrm{softmax}\!\Big(\tfrac{Q^{P}_i(K^{D}_i)^{\top}}{\sqrt{d_h}}\Big)V^{D}_i,
  \end{aligned}
\end{equation}
with head dimension $d_h$. Heads are concatenated and projected to $H_D,H_P$.
We use $d_e=384$, $h=4$ and $d_h=128$ ($\approx$1.6M attention parameters).
Contextual and attention vectors are fused by an equal-weight residual
combination:
\begin{equation}
  D_f=0.5\,H_D+0.5\,D,\qquad P_f=0.5\,H_P+0.5\,P.
\end{equation}
This fixed fusion keeps pretrained information while injecting cross-sequence context; alternative weights or a learnable gate are not evaluated here and are noted as a limitation.

\subsection{Convolutional Classification and Objective}
Two independent 1D CNN branches process $D_f$ and $P_f$. Each branch applies two
1D convolutions (32 channels, kernel size 5), each followed by ReLU, max pooling
and dropout. The drug and protein feature vectors are concatenated into a
4960-dimensional vector and passed to a three-layer MLP
($4960\!\to\!256\!\to\!32\!\to\!1$) with ReLU, dropout $0.1$ and a final sigmoid
returning $\hat{y}=F(d,p)$. The reported configuration uses
SmoothL1Loss~\cite{girshick2015fast} for robustness to large residuals; because
DTI is a classification task with class imbalance, results are interpreted with
ROC-AUC, PR-AUC, sensitivity and specificity rather than the loss alone.

\section{Experiments}
\subsection{Datasets, Protocol and Metrics}
We use BIOSNAP, DAVIS and BindingDB with the benchmark partitions reported for
MolTrans and the pretrained BERT baseline~\cite{huang2021moltrans,kang2022finetuning}.
BIOSNAP is built from known interactions balanced with sampled negatives; for
DAVIS and BindingDB, pairs with $K_d<30$ are treated as positive, leaving
strongly imbalanced test sets. BIOSNAP has 4{,}510 drugs / 2{,}181 proteins,
DAVIS 68 / 379, and BindingDB 10{,}665 / 1{,}413. The pair counts used for
the main comparison are the original benchmark partitions: BIOSNAP
9619/9619 training, 1374/1374 validation and 2748/2748 test positives/negatives;
DAVIS 1043/1043, 160/2846 and 303/5708; BindingDB 6334/6334, 927/5717 and
1905/11384. 
Drug sequences are truncated or padded to 100 tokens and
proteins to 545. These thresholds were selected from the observed length
distributions in the experimental report and cover approximately 95\% of the
benchmark sequences; larger protein cutoffs were tested during development but
reduced feasible batch sizes because of GPU memory. Training uses Adam
(learning rate $5\times10^{-5}$), batch size 36, dropout $0.1$ and up to 35
epochs, with PyTorch, PyTorch Lightning and Hugging Face Transformers on four
NVIDIA P100 GPUs. BERT4DTI results are
averaged over five runs; per-run model selection uses validation ROC-AUC.
The hyperparameter search considered learning rates
$\{10^{-6},10^{-5},5\times10^{-5},10^{-4}\}$, dropout values
$\{0.1,0.2,0.3\}$ and batch sizes $\{32,36\}$. We kept the fixed benchmark
splits to remain comparable with published baselines.
We report ROC-AUC, PR-AUC, sensitivity and specificity; PR-AUC and
specificity are essential context for the imbalanced DAVIS/BindingDB settings,
so a sensitivity gain is not read as uniform superiority. For additional
operating-point context, precision, F1 and MCC are reconstructed for BERT4DTI
from the reported test distributions and mean sensitivity/specificity: 0.823,
0.851 and 0.694 on BIOSNAP; 0.217, 0.349 and 0.389 on DAVIS; and 0.416, 0.572
and 0.527 on BindingDB. These reconstructed values should be treated as
approximate because they are derived from aggregate rates rather than from
per-seed confusion matrices. We report trainable-parameter counts.

\begin{table*}[t!]
  \centering
  \caption{Performance comparison on BIOSNAP, DAVIS and BindingDB. Best value in
    each dataset/metric group is in \textbf{bold}. Baseline values are taken from
    the corresponding benchmark reports.}
  \label{tab:main}
  \small
  \begin{tabular}{llcccc}
    \toprule
    Dataset & Method & ROC-AUC & PR-AUC & Sensitivity & Specificity \\
    \midrule
    \multirow{6}{*}{BIOSNAP}
      & MolTrans         & $0.895\pm0.002$ & $0.901\pm0.004$ & $0.775\pm0.032$ & $0.851\pm0.014$ \\
      & DeepDTA          & $0.876\pm0.005$ & $0.883\pm0.006$ & $0.781\pm0.015$ & $0.824\pm0.012$ \\
      & Fine-tuned BERT  & $0.914\pm0.006$ & $0.900\pm0.007$ & $0.862\pm0.025$ & $0.847\pm0.007$ \\
      & DLM-DTI          & $0.914\pm0.003$ & $0.914\pm0.006$ & $0.848\pm0.016$ & $0.844\pm0.024$ \\
      & ICAN             & $0.871$ & $0.886$ & $0.799$ & $0.786$ \\
      & \textbf{BERT4DTI}& $\mathbf{0.920\pm0.002}$ & $\mathbf{0.924\pm0.003}$ & $\mathbf{0.882\pm0.013}$ & $0.810\pm0.020$ \\
    \midrule
    \multirow{6}{*}{DAVIS}
      & MolTrans         & $0.907\pm0.002$ & $\mathbf{0.404\pm0.016}$ & $0.800\pm0.022$ & $0.876\pm0.013$ \\
      & DeepDTA          & $0.880\pm0.007$ & $0.302\pm0.044$ & $0.764\pm0.045$ & $0.865\pm0.020$ \\
      & Fine-tuned BERT  & $\mathbf{0.920\pm0.002}$ & $0.395\pm0.007$ & $0.824\pm0.026$ & $\mathbf{0.889\pm0.015}$ \\
      & DLM-DTI          & $0.895\pm0.003$ & $0.373\pm0.017$ & $0.833\pm0.044$ & $0.802\pm0.070$ \\
      & ICAN             & $0.903$ & $0.372$ & $0.884$ & $0.766$ \\
      & \textbf{BERT4DTI}& $\mathbf{0.920\pm0.005}$ & $0.370\pm0.022$ & $\mathbf{0.891\pm0.016}$ & $0.829\pm0.018$ \\
    \midrule
    \multirow{6}{*}{BindingDB}
      & MolTrans         & $0.914\pm0.001$ & $0.622\pm0.007$ & $0.797\pm0.005$ & $0.896\pm0.007$ \\
      & DeepDTA          & $0.913\pm0.003$ & $0.622\pm0.012$ & $0.780\pm0.035$ & $0.915\pm0.016$ \\
      & Fine-tuned BERT  & $\mathbf{0.922\pm0.001}$ & $0.623\pm0.010$ & $0.814\pm0.025$ & $\mathbf{0.916\pm0.016}$ \\
      & DLM-DTI          & $0.912\pm0.004$ & $\mathbf{0.643\pm0.006}$ & $0.888\pm0.014$ & $0.793\pm0.015$ \\
      & ICAN             & $0.900$ & $0.604$ & $0.846$ & $0.815$ \\
      & \textbf{BERT4DTI}& $0.912\pm0.003$ & $0.630\pm0.004$ & $\mathbf{0.912\pm0.005}$ & $0.786\pm0.015$ \\
    \bottomrule
  \end{tabular}
\end{table*}

\subsection{Main Benchmark Results}
Table~\ref{tab:main} reports the comparison. On BIOSNAP, BERT4DTI obtains the
best ROC-AUC ($0.920$) and PR-AUC ($0.924$) and the highest sensitivity, while
MolTrans and the BERT baselines reach higher specificity: the model retrieves
positive pairs well on a balanced test set but admits more false positives. On
DAVIS, BERT4DTI matches the best ROC-AUC ($0.920$) and gives the highest
sensitivity ($0.891$), whereas MolTrans has the best PR-AUC and fine-tuned BERT
the best specificity; the strongly imbalanced test distribution makes PR-AUC and specificity essential context. On BindingDB, BERT4DTI again gives the highest sensitivity ($0.912$), with full BERT fine-tuning best on ROC-AUC/specificity and DLM-DTI best on PR-AUC. BERT4DTI is therefore best viewed as a competitive model with a pronounced positive-retrieval profile rather than a uniformly superior method. 

\subsection{Ablation and Parameter Efficiency}
We ablate the CNN and mutual-attention modules individually and jointly on
DAVIS (Table~\ref{tab:ablation}). Removing mutual attention lowers ROC-AUC from $0.92$ to $0.89$, PR-AUC from $0.37$ to $0.29$ and specificity from $0.83$ to $0.75$, indicating that attention contributes notably to PR-AUC under this setting. Removing both modules recovers PR-AUC close to the full model but yields a larger trainable size (183M vs.\ 125M). Table~\ref{tab:params} compares trainable parameters: BERT4DTI uses
125M, far fewer than 353M for full BERT fine-tuning, though more than MolTrans
(62.8M) and DLM-DTI (86.7M); it occupies an intermediate position, retaining
more adaptation capacity than lighter alternatives while avoiding the full size
of the paired encoders. A parameter count alone cannot establish wall-clock
efficiency, which depends on sequence length, frozen-layer computation,
attention memory, batch size and hardware. Accordingly, we report
BERT4DTI as parameter-efficient relative to full BERT fine-tuning.

Additional analyses in the experimental study (reported descriptively, not as
controlled comparisons) include a cold-split evaluation holding out 20\% of
drugs or proteins (best held-out ROC-AUC $0.901$ for unseen drugs), a
limited-data setting where BERT4DTI keeps the highest ROC-AUC down to $0.770$
with 95\% of training data removed, a BIOSNAP protein-family stratification
(strongest on GPCRs and ion channels, PR-AUC $0.99$), and an exploratory
positive-pair study on 13 CYP interactions (average score $0.727$ vs.\ $0.479$
for MolTrans). These analyses contain no negative pairs or similarity-clustered
splits and do not support claims of general external generalization.

\begin{table}[t!]
  \centering
  \caption{Ablation on DAVIS.}
  \label{tab:ablation}
  \small
  \begin{tabular}{lcccc}
    \toprule
    Configuration & ROC-AUC & PR-AUC & Sens. & Spec. \\
    \midrule
    BERT4DTI                    & $0.92$ & $0.37$ & $0.89$ & $0.83$ \\
    \quad w/o CNN               & $0.91$ & $0.36$ & $0.91$ & $0.76$ \\
    \quad w/o mutual attention  & $0.89$ & $0.29$ & $0.90$ & $0.75$ \\
    \quad w/o CNN \& attention  & $0.91$ & $0.37$ & $0.90$ & $0.80$ \\
    \bottomrule
  \end{tabular}
\end{table}

\begin{table}[t!]
  \centering
  \caption{Reported trainable parameter counts.}
  \label{tab:params}
  \small
  \begin{tabular}{lc}
    \toprule
    Model & Trainable parameters (M) \\
    \midrule
    MolTrans                              & 62.8 \\
    DLM-DTI                               & 86.7 \\
    BERT4DTI                              & 125.0 \\
    BERT4DTI w/o CNN \& attention         & 183.0 \\
    Full BERT fine-tuning                 & 353.0 \\
    \bottomrule
  \end{tabular}
\end{table}

\subsection{Discussion}
Table~\ref{tab:main} reports the benchmark comparison on BIOSNAP, DAVIS and
BindingDB. On BIOSNAP, BERT4DTI achieves the best ROC-AUC ($0.920$) and PR-AUC
($0.924$), together with the highest sensitivity. This indicates that the
combination of pretrained sequence encoders and mutual attention is effective on
a balanced benchmark where positive and negative pairs are equally represented.
However, its specificity remains lower than that of MolTrans and fine-tuned BERT,
showing that BERT4DTI favours positive-pair retrieval over conservative rejection
of negatives.

This positive-retrieval profile is also observed on DAVIS and BindingDB. On
DAVIS, BERT4DTI matches the best ROC-AUC and obtains the highest sensitivity,
whereas MolTrans obtains the best PR-AUC and fine-tuned BERT the best
specificity. On BindingDB, BERT4DTI again achieves the highest sensitivity, while
full BERT fine-tuning gives the best ROC-AUC and specificity and DLM-DTI gives
the best PR-AUC. These results suggest that BERT4DTI is well suited to early
screening scenarios where recovering potential interactions is important, but
they also show that it should not be interpreted as uniformly superior across all
operating criteria. In imbalanced screening, PR-AUC and specificity remain
important because false positives increase the experimental follow-up burden.

Table~\ref{tab:ablation} supports the contribution of the mutual-attention
module. Removing mutual attention decreases ROC-AUC from $0.92$ to $0.89$,
PR-AUC from $0.37$ to $0.29$, and specificity from $0.83$ to $0.75$ on DAVIS.
This suggests that cross-sequence token alignment provides information beyond
independent pretrained encodings followed by late fusion. Removing the CNN branch
has a smaller effect on ROC-AUC and PR-AUC but still reduces specificity,
indicating that local convolutional aggregation remains useful after mutual
attention.

The parameter analysis in Table~\ref{tab:params} positions BERT4DTI between
lighter sequence models and full BERT fine-tuning. With 125M trainable
parameters, BERT4DTI is substantially smaller than full BERT fine-tuning
(353M), while preserving more adaptation capacity than MolTrans and DLM-DTI.
Therefore, the efficiency claim concerns trainable model size, not measured
runtime. Runtime, peak memory and inference latency depend on sequence length,
frozen-layer computation, batch size and hardware, and should be evaluated in a
dedicated deployment study.

Additional analyses, including held-out drug/protein settings, scarce-data
experiments, protein-family stratification and the CYP positive-pair case study,
provide complementary evidence that BERT4DTI preserves useful ranking behaviour
beyond the main random-pair benchmark. These analyses remain descriptive because
they do not replace similarity-clustered splits, duplicate-pair auditing,
probability calibration or an external benchmark with both positive and negative
evidence. Therfore, this  comparison relies on the original
benchmark partitions, and merged-dataset or SMOTE-based variants are not used as
central evidence.

\section{Conclusion}
BERT4DTI combines ChemBERTa and ProtBERT with partial fine-tuning,
bidirectional mutual attention and convolutional classification for
sequence-based DTI prediction. The model obtains strong results across three
benchmarks, especially in terms of positive-pair retrieval, and the ablation
study supports the value of mutual attention for cross-sequence interaction
modelling. At the same time, the results reveal a clear trade-off: on imbalanced
datasets, competing methods may obtain better PR-AUC or specificity. BERT4DTI
should therefore be viewed as a competitive screening model with reduced
trainable size relative to full BERT fine-tuning, rather than as a universally
dominant predictor.
Future work will extend BERT4DTI toward knowledge-grounded DTI prediction by integrating biomedical knowledge graphs and retrieval-augmented language models to support explainable, evidence-based interaction prioritization.

\section*{Acknowledgment}
This research was supported by the Natural Sciences and Engineering Research Council of Canada (NSERC) through a Discovery Grant (RGPIN-2025-06792) and internal funding from TÉLUQ University (FAR3 program).

\section*{GenAI Usage Disclosure}
Generative AI tools were used only for language polishing and formatting
assistance. All scientific contributions, experimental results, interpretations
and conclusions are the responsibility of the authors and were checked against
the original experimental report.

\end{document}